\documentclass[conference]{IEEEtran}
\usepackage{times}
\usepackage{xcolor}
\usepackage{multirow}
\usepackage{hyperref}
\usepackage{graphicx} 
\usepackage{amsmath}
\usepackage{siunitx}
\usepackage{booktabs}
\usepackage[T1]{fontenc}
\usepackage{array}
\usepackage{booktabs}
\ifCLASSINFOpdf
\else
\fi
\usepackage{amsmath}
\usepackage{algorithmic}

\usepackage{array}
\begin{document}

%
\title{LOLgorithm: Integrating Semantic,Syntactic and Contextual Elements for Humor Classification}

\author{
\IEEEauthorblockN{Tanisha Khurana\IEEEauthorrefmark{4}\IEEEauthorrefmark{1},
Vikram Pande\IEEEauthorrefmark{4}\IEEEauthorrefmark{1},
Kaushik Pillalamarri\IEEEauthorrefmark{2}\IEEEauthorrefmark{1}, 
Munindar Singh\IEEEauthorrefmark{2}}
\IEEEauthorblockA{\IEEEauthorrefmark{4}Department of Electrical Engineering at North Carolina State University, \{tkhuran3, vspande\}@ncsu.edu}
\IEEEauthorblockA{\IEEEauthorrefmark{2}Department of Computer Science at North Carolina State University, \{spillal2, mpsingh\}@ncsu.edu}
\IEEEauthorblockA{\IEEEauthorrefmark{1}\textsuperscript{joint first authors}}
}


%


\maketitle



%
\IEEEpeerreviewmaketitle

\begin{abstract}

This paper explores humor detection through a linguistic lens, prioritizing syntactic, semantic, and contextual features over computational methods in Natural Language Processing. We categorize features into syntactic, semantic, and contextual dimensions, including lexicons, structural statistics, Word2Vec, WordNet, and phonetic style. Our proposed model, Colbert, utilizes BERT embeddings and parallel hidden layers to capture sentence congruity. By combining syntactic, semantic, and contextual features, we train Colbert for humor detection. Feature engineering examines essential syntactic and semantic features alongside BERT embeddings. SHAP interpretations and decision trees identify influential features, revealing that a holistic approach improves humor detection accuracy on unseen data. Integrating linguistic cues from different dimensions enhances the model's ability to understand humor complexity beyond traditional computational methods.

\end{abstract}

\section{Problem Description}
Humor is a fascinating and puzzling area of study in the field of computers understanding human language. In recent times, computers have transformed from being simple machines that follow instructions to becoming smart agents that interact with people and learn from them. When a computer talks to a person, if it can understand and appreciate the humor in what the person is saying, it can grasp the real meaning of human language better. This, in turn, helps the computer make better decisions to enhance the user's experience. So, working on techniques that allow computers to "get" humor in human conversations and adjust their responses accordingly is something we should pay special attention to.

Humor Recognition is about figuring out if a sentence is funny in a given situation. It's a tough problem in understanding human language. First, humor is tricky because different people may find different things funny even in the same sentence. Second, humor often depends on knowing a lot about the context. For example, think about these sentences: "The one who invented the door knocker got a No Bell prize" and "Veni, Vidi, Visa: I came, I saw, I did a little shopping." To find them funny, you need to know a bit about culture and language. Lastly, there are many types of humor, like wordplay, irony, and sarcasm, but we don't have a clear way to categorize them all. So, it's almost impossible to create a computer program that can recognize all types of humor, just like humans can't always classify them perfectly either.

In this work, we formulate humor recognition as a classification task in which we distinguish between humorous and non-humorous instances.
Exploring the syntactical structure involves leveraging Lexicons to capture sentiment counts within a sentence, while Statistics of Structural Elements (SSE) \cite{liu2018exploiting} encapsulates the statistical insights of Noun phrases, Word phrases, and more. Unveiling the semantic layers of humor delves into Word2Vec \cite{mikolov2013efficient} embeddings , analyzing incongruity, ambiguity, and phonetic structures within sentences. Additionally, contextual information is harnessed through ColBERT embeddings \cite{annamoradnejad2020colbert}. For each latent structure, we design a set of
features to capture the potential indicators of humor.

\subsection{Example Scenario}
Imagine you work for a social media analytics company, and one of your clients is interested in understanding and quantifying humor within user-generated content on their platform. They want to identify and categorize humorous posts, as well as analyze the types of humor that resonate the most with their audience. Humor classification will be extremely beneficial for these scenarios. 

Amazon for example employs humor detection in product question-answering systems \cite{ziser2020humor} as some products attract humor due to their unreasonable price and their peculiar functionality. Detecting humorous questions in such systems is important for sellers, to better understand user engagement with their products. It is also important to inform users about the flippancy of humorous questions, and that answers for such questions should be taken with a grain of salt.

\section{Relevant Literature}
Many of the exciting studies on humor recognition are formulated as binary classification problems and try to identify jokes using a set of linguistic features. For example, Zhang et al. \cite{zhang2014recognizing} developed various humor-related features based on humor theories, linguistic rules, and emotions. They used these features in a Gradient Boosting Regression Tree model to recognize humor. Yang et al. \cite{yang2015humor} tried to identify several semantic structures behind humor and design sets of features for each structure, they extracted anchors that enable humor in the sentence and then used a computational approach to recognize humor.  They explored the types of latent semantic structure behind humor namely incongruity, ambiguity, interpersonal affect, and phonetic style. Liu et al. \cite{liu2018exploiting} extended this research by adding syntactical structures for humor detection such as statistics of structural elements, production rules, and dependency relations.
Annamoradnejad et al. \cite{annamoradnejad2020colbert} have leveraged the power of the transformer's embeddings to capture the contextual information from a given sentence.

\subsection{DATASET}
We employ two distinct datasets, for training and testing purposes:

ColBERT Dataset: The ColBERT dataset \cite{colbertdataset} has 200,000 statements with humor labels and is accessible on Kaggle for model training and evaluation. We partition it into training and validation sets.

For the unseen test set, we scraped funny one-liners from the Bestlife website to gather jokes. We preprocess the data by removing punctuation and lemmas and manually labeling the jokes as humorous or not for unseen datasets for model evaluation.
Also, we've obtained around 2K samples of sentences from Reddit, out of which some are funny and some are not, which would also be used for further evaluating the model's performance on unseen data.

\subsection{CODE}
We use libraries such as PyTorch \cite{paszke2019pytorchimperativestylehighperformance}, TensorFlow, Hugging-Face \cite{huggingface}, NLTK \cite{bird2009natural}, SpaCy \cite{spacy2}, NRCLex \cite{nrclex}, Sklearn \cite{scikit-learn}, and some inherent Python libraries to help in the coding part.

\section{Problem importance}
Humor classification is a crucial problem and our solution takes into account the contextual, semantic, and syntactic meaning of the joke. Our problem has various applications in customer support and chatbot development. Many companies utilize conversational services
to recommend items and counsel consumers. There is also a 
growing demand for emotional chatterbots for user's higher satisfaction in various commercial fields. 
Siri could become more ‘human’ if she had the ability to recognize social cues like humor and respond to them with laughter. As described in our example scenario, our problem will be useful in social media analytics and for question-answering systems for products on Amazon\cite{ziser2020humor}. Our problem also has further usage when combined with speech as well. For example, sitcoms are written and performed primarily to cause laughter, so the ability to detect precisely why something will draw laughter is of extreme importance to screenwriters and actors \cite{purandare2006humor}.

\section{Methodology}
The aim of the project was to work more with linguistic features rather than exploring computational methods in the area of Natural Language Processing. For the features we chose to work with, we've categorized them into three types for experimentation: syntactic, semantic, and contextual features. Syntactic features help understand how words function structurally and influence model predictions. Semantic analysis delves into word and sentence meanings. Exploring contextual dependencies within sentences is also crucial. We're generating these features using methods described later in this section and analyzing these features to identify the most influential ones driving model predictions using basic SHAP interpretations \cite{lundberg2017unified}, and visualizing decision trees \cite{breiman1986classification}. We then model these features independently and together by computational methods.

\subsection{\textbf{Syntactical information}}
\subsubsection{\textbf{Lexicons}}
	In some cases, individual words can be inherently funny. Therefore, if a sentence contains amusing words, the entire sentence may be humorous. With this assumption in mind, our initial approach involves extracting syntactical information using lexicons \cite{jurafsky_speech_2024} \cite{mohammad2010emotions}. Specifically, we employ the NRC word emotion lexicons for this purpose. 
	Utilizing this lexicon, we represent words as vectors, where each entry signifies the word's similarity to various emotions such as anger, anticipation, disgust, fear, joy, sadness, and surprise, as well as its similarity to the general sentiments of positivity and negativity. By employing these word representations, we aim to capture the syntactical information embedded within each word.
	To obtain syntactical embeddings for the entire sentence, we aggregate the vectors of each word in the sentence, thereby encapsulating the overall syntactical characteristics of the sentence. We make use of the available NRCLex library to measure the emotional affect of a body of jokes. We get the scores for emotions such as \textit{fear, anger, anticipation, trust, surprise, positive, negative, sadness, disgust}, and \textit{joy} for each joke. 

\begin{figure}[ht]
    \centering
    \includegraphics[width=0.5\textwidth]{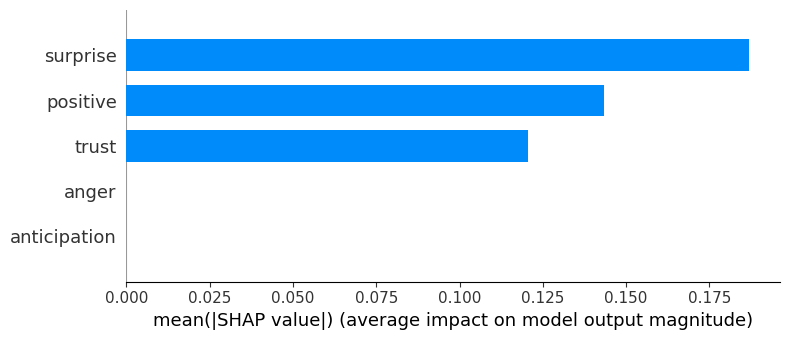}
    \caption{SHAP on NRCLex}
    \label{fig:shap2}
\end{figure}
\begin{figure}[ht]
    \centering
    \includegraphics[width=0.5\textwidth]{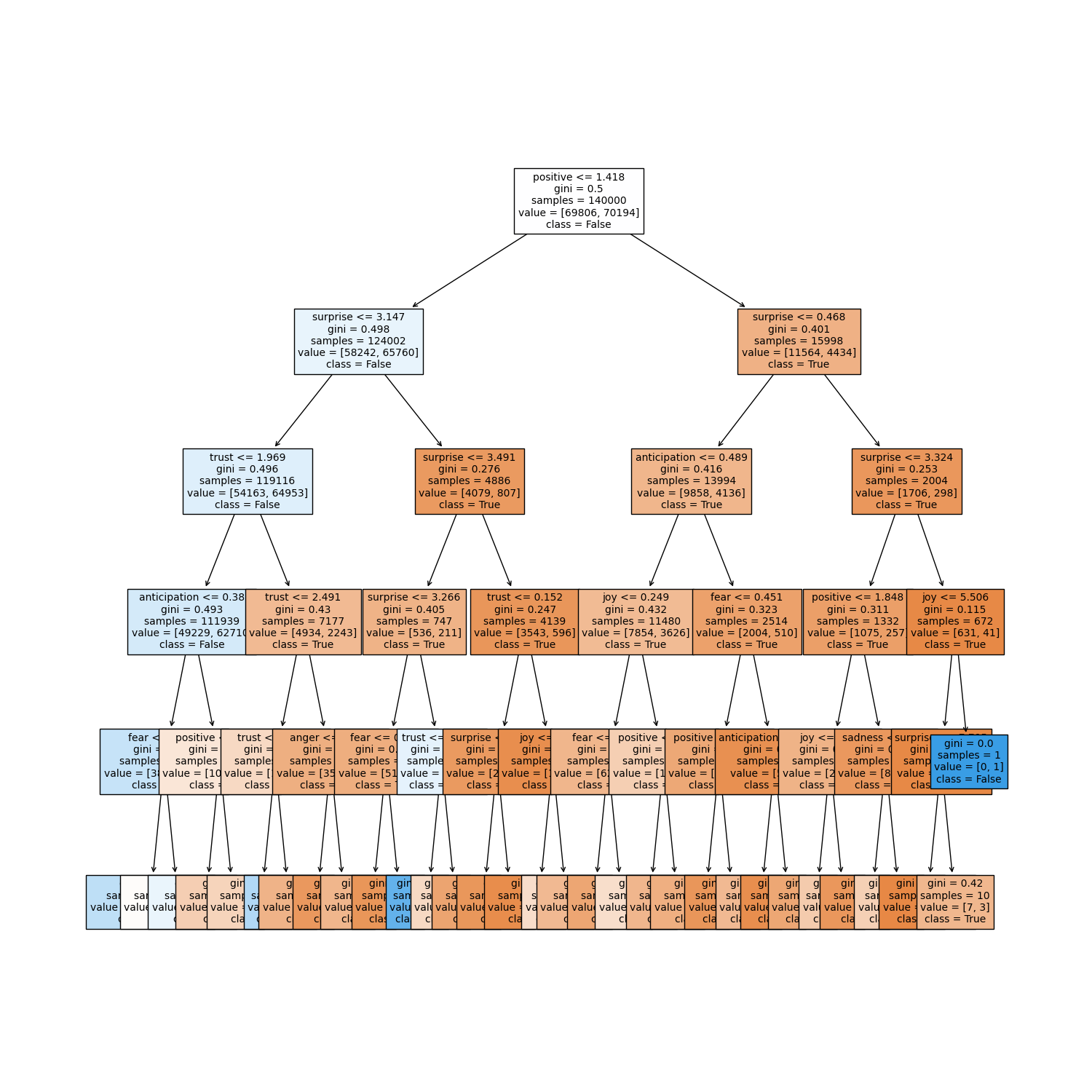}
    \caption{DT on NRCLex}
    \label{fig:shap3}
\end{figure}
\subsubsection{\textbf{Statistics of Structural Elements}}
	As described by Liu et al.\cite{liu2018exploiting} we follow the methods to acquire syntactic structures for humor recognition. Among these methods, we intend to focus on utilizing statistics related to structural elements to generate features for subsequent classification. These features encompass complexity metrics, phrase length ratio, average phrase length, and more. Statistics of structural elements have been effective in evaluating the linguistic quality of text \cite{nenkova}. We implement the following syntactic features: 

 \begin{itemize}
     \item {\textbf{Complexity Metrics}: Complexity metrics can be calculated by measuring the differences from the perspective of sentence complexity as humorous and non-humorous texts may differ in the way they express intentions.  Therefore, the number of noun phrases (NP count), verb phrases(VP count), prepositional phrases (PP count), and subordinating conjunctions (SBAR count) are counted respectively as features.}
     
     \item {\textbf{Phrase Length Ratio}: The length ratio (LR) for PP, NP, and VP is individually calculated. This ratio represents the number of words in each phrase type divided by the total sentence length. }
     
     \item  {\textbf{Avergae Phrase Length}: The average phrase length is determined by dividing the total number of words within each phrase by the respective number of phrase types. It's essential to note two distinct calculation methods: one (AP L1) accounts for nested phrases. For instance, in a VP phrase (VP1...(...VP2...)), the VP's length equals the sum of the lengths of VP1 and VP2. The other approach (AP L2) considers only the maximum phrase length, where the phrase length is determined by the length of VP1.}
     
     \item{\textbf{Ratio of PP or NP within a VP (RP NV)}: If a VP encompasses NP or PP, this value corresponds to the average length of NP or PP divided by the length of VP.}
 \end{itemize}

 We make use of the NLTK package to obtain Statistics of Structural Elements and these features can be called SSE features (SSE).

\begin{figure}[htbp]
    \centering
    \includegraphics[width=0.5\textwidth]{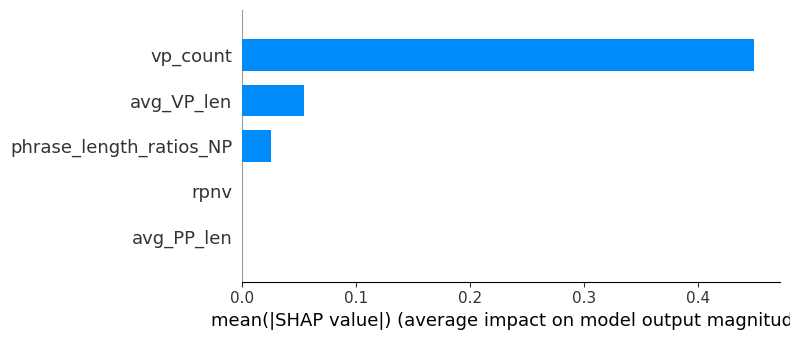}
    \caption{SHAP on Syntactic Features}
    \label{fig:shap4}
\end{figure}
\begin{figure}[htbp]
    \centering
    \includegraphics[width=0.5\textwidth]{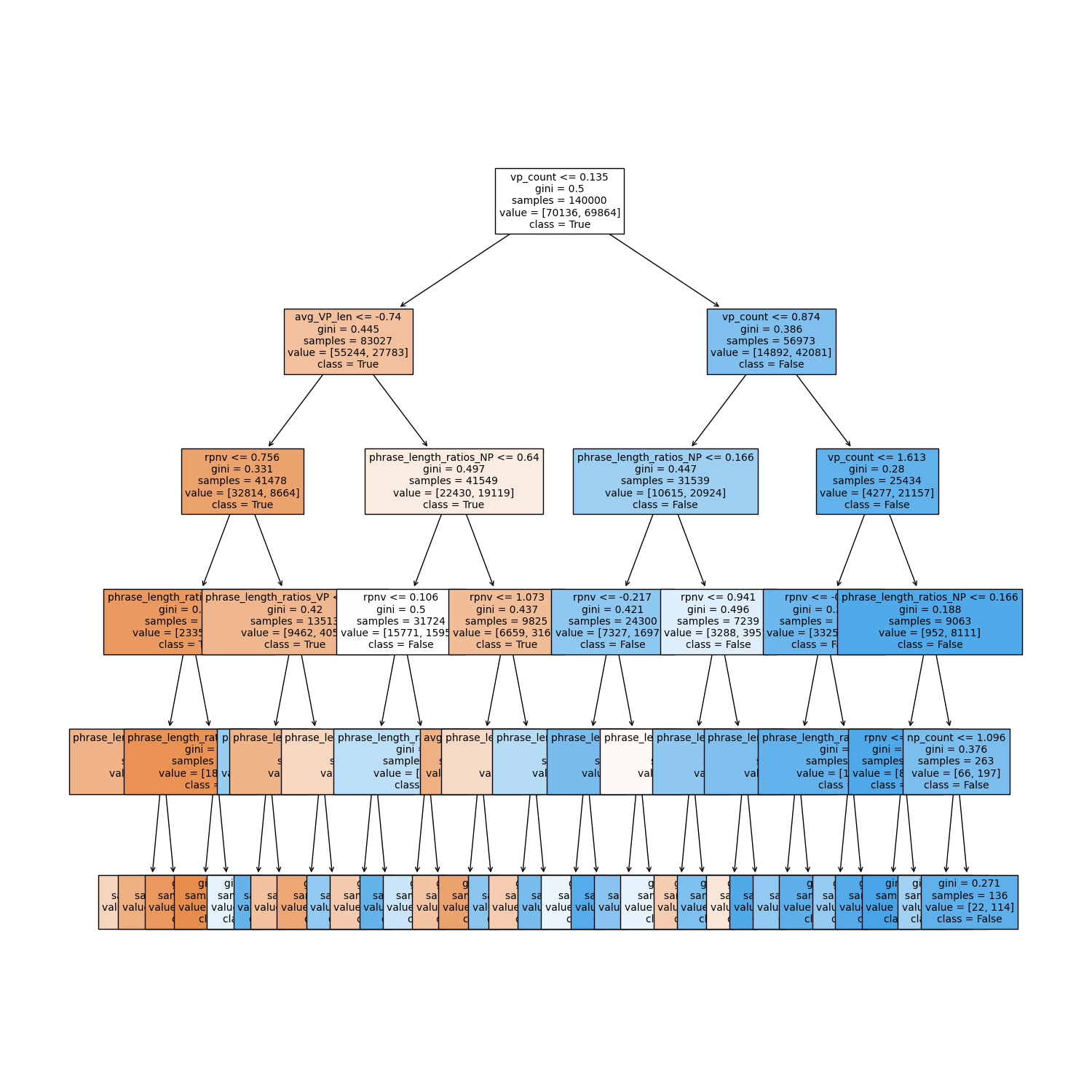}
    \caption{DT on Syntactic Features}
    \label{fig:dt2}
\end{figure}
\subsection{\textbf{Semantic information}}
	Among the 12 humor theories \cite{hempelmann2008computational}, also known as humor structures, \cite{yang2015humor} we explore the latent semantic structures behind humor in three aspects: (a) Incongruity, (b) Ambiguity (c) Phonetic Style.

\begin{itemize}
    \item{\textbf{Incongruity Structure}} \\
Laughter often arises from the union of seemingly contradictory or incongruous elements, forming a unique relationship or assemblage within the mind (Lefcourt, 2001) \cite{lefcourt2002humor}. The crux of humor lies in the incongruity, the separation of one idea from another (Paulos, 2008) \cite{paulos}. Humor often thrives on specific types of incongruity, like opposition or contradiction. We extract two types of features to assess the meaning distance between pairs of content words in a sentence by leveraging Word2Vec \cite{mikolov2013efficient} to gauge the semantic connections within a sentence. We describe incongruity through the following two features:

\subitem{\textbf{Disconnection:} Representing the maximum meaning (semantic) distance among word pairs in a sentence.}
\subitem{\textbf{Repetition:} Representing the minimum meaning (semantic) distance among word pairs in a sentence.} \\ 
   
    \item{\textbf{Ambiguity Theory}} \\
Humor and ambiguity often come together when the listener tries to interpret the meaning. Ambiguity arises when the words in a sentence can be grouped in multiple ways, resulting in various underlying interpretations. For example: \\
\textit{I saw the man on the hill with the telescope.} \\
Different possible meanings of words lead to diverse understandings for readers. To capture this ambiguity within a sentence, we employ WordNet and assess it as follows: 
\subitem{\textbf{Sense Combination}: This computation involves identifying Nouns, Verbs, Adjectives, and Adverbs through a POS tagger \cite{bird2009natural}. Subsequently, we consider the potential meanings of these words \textit{(w1, w2, ... ,wk)}
 via WordNet \cite{Fellbaum1998}, calculating the sense combinations as 
 
\begin{equation} label{eu_eqn}
    \log\left(\prod_{\text{sense} \in \text{senses}} \pi(\text{sense})\right)
\end{equation}

\subitem{\textbf{Sense Farmost}: the largest Path Similarity of any word senses in a sentence.}
\subitem{\textbf{Sense Closest}: the smallest Path Similarity of any word senses in a sentence.} \\ 
    \item{\textbf{Phonetic Style}} \\
Mihalcea et al. \cite{mihalcea2005making} suggests that the phonetic characteristics of humorous sentences hold significant importance alongside their content. Many one-liners exhibit linguistic elements like alliteration, word repetition, and rhyme, creating a comedic effect regardless of whether the joke is actually funny. An alliteration chain involves consecutive words starting with the same sound, while a rhyme chain comprises words ending with the same syllable. To extract these phonetic features, we utilize the CMU Pronouncing Dictionary \cite{cmudict} and create four features:
\subitem{\textbf{Alliteration}: quantifies the count and maximum length of alliteration chains within a sentence.}
\subitem{\textbf{Rhyme}: measures the count and maximum length of rhyme chains.}}

\end{itemize} 
 
These can be called Human Theory Driven Features (HTF).

\begin{figure}[ht]
    \centering
    \includegraphics[width=0.45\textwidth]{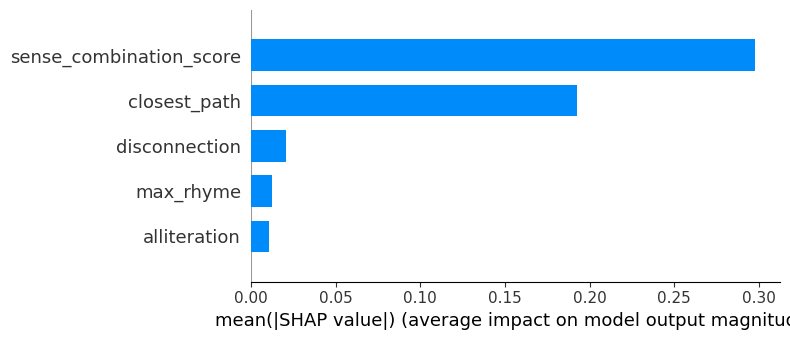}
    \caption{SHAP on semantic features}
    \label{fig:semantic}
\end{figure}
\begin{figure}[ht]
    \centering
    \includegraphics[width=0.5\textwidth]{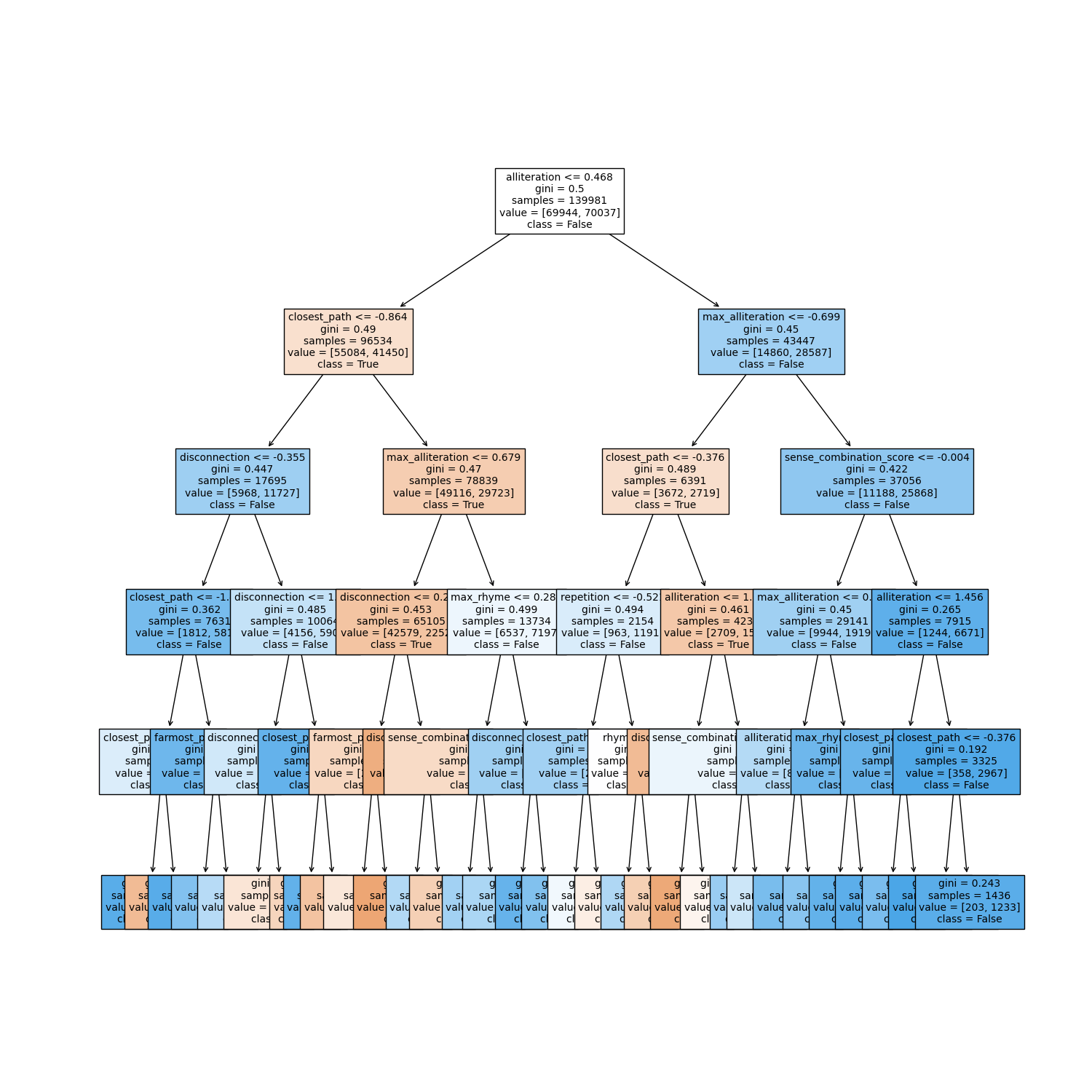}
    \caption{DT on semantic features}
    \label{fig:dt3}
\end{figure}

\subsection{\textbf{Contextual information}}
By looking at the general structure of a joke to understand the underlying linguistic features that make a text laughable, Many suggested that humor arises from the sudden transformation of an expectation into nothing. In this way, the punchline, as the last part of a joke, destroys the perceiver’s previous expectations and brings humor to its incongruity. The punchline is related to previous sentences but is included in opposition to previous lines in order to transform the reader’s expectation of the context.

The proposed method for humor detection focuses on the structure of humor in text. It observes that individual sentences in a joke may seem normal and non-humorous when read separately but become humorous when considered together in context. To capture this, the method employs the model as described by Colbert \cite{annamoradnejad2020colbert} a neural network architecture with separate paths for sentence-level and whole-text features. The process involves tokenizing and encoding sentences using BERT sentence embeddings \cite{devlin2018bert}, followed by parallel hidden layers to extract mid-level features for each sentence. \newline

Each sentence segment is given max\_sentence\_length as 20 tokens and a maximum of 5 segments are considered for an individual joke. For the whole sentence, 100 tokens are considered.  The model aims to detect relationships between sentences, especially the punchline's connection to the rest, and examines word-level connections in the entire text. \newline

\begin{figure*}[ht]
    \centering
    \includegraphics[width=0.9\textwidth]{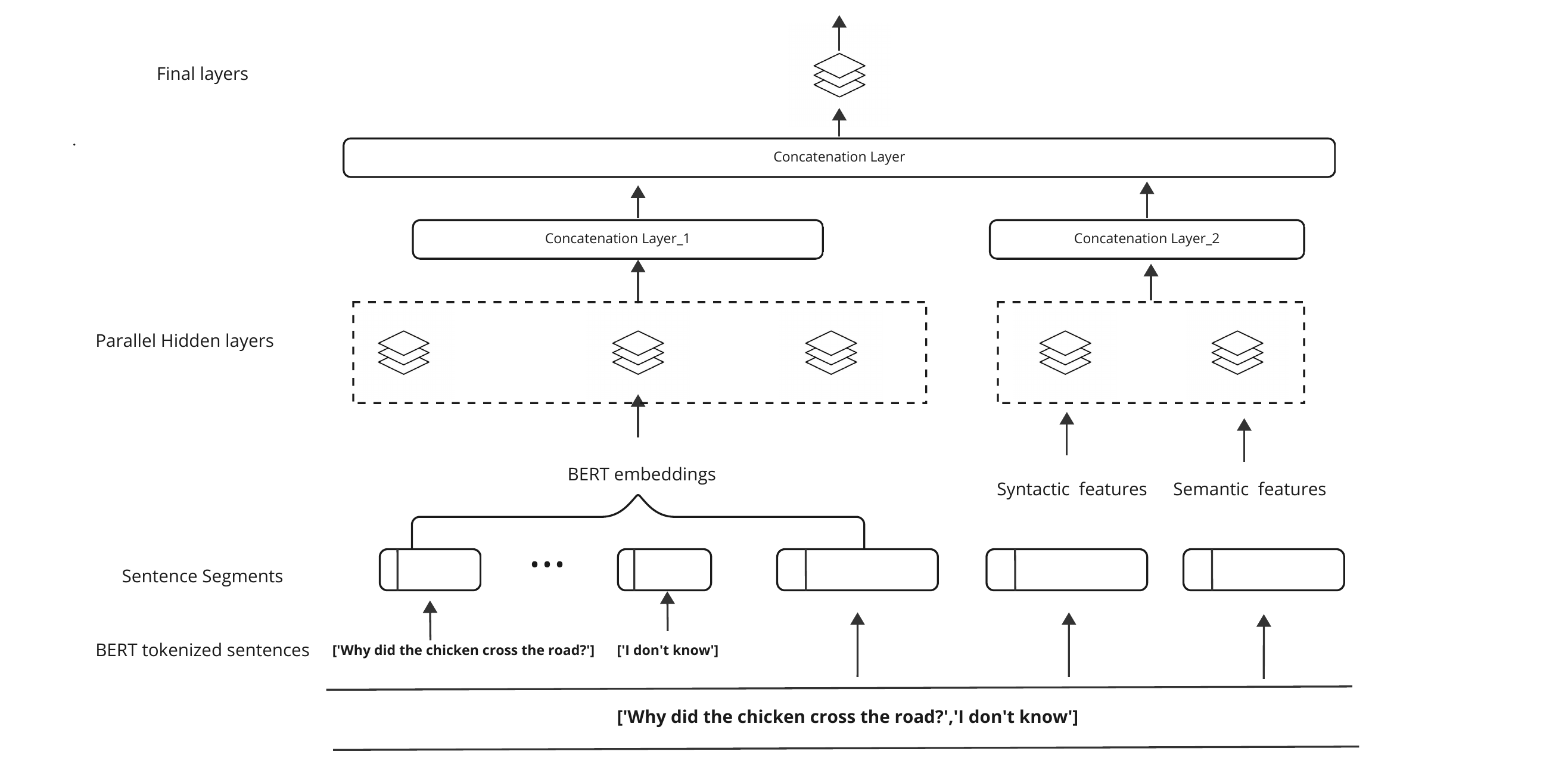}
    \caption{Colbert Architecture with Syntactic and Semantic features}
    \label{fig:colbert}
\end{figure*}

These BERT sentence embeddings \cite{devlin2018bert} are generated by inputting these tokens into the BERT model, resulting in vectors of size 768. The model involves eight neural network layers, with each sentence processed through a separate parallel line of three hidden layers. These layers are concatenated in the fourth layer and continue sequentially to predict a single target value. The implementation utilizes the huggingface \cite{huggingface} library for BERT and keras.tensorflow \cite{tensorflow2015} for the neural network. Notably, training occurs solely on the neural network, not the BERT model. The chosen BERT variant is the smaller-sized BERTBASE, with 12 layers, 768 hidden states, 12 heads, and 110 million parameters, pre-trained on lower-cased English text from BooksCorpus \cite{zhu2015books} and English Wikipedia.
The final layers of the neural network integrate the outputs from all paths to predict the congruity of sentences and identify shifts in the reader's viewpoint after encountering the punchline.

Parallel hidden layers in a neural network process the BERT embeddings \cite{devlin2018bert} for each sentence, extracting mid-level features related to context and sentence type, resulting in a 20-dimensional vector for each sentence.

\subsection{\textbf{Combination of Syntactic, Semantic with ColBERT}}
In our modifications we firstly combined all the previously described syntactical, semantic, and contextual information obtained from NRC, HTF into one feature list of 33 features for 200k inputs. These features were given to parallel hidden layers and then concatenated with the Bert embeddings obtained from Colbert data preprocessing passed through the Dense neural network layers. The model is finalized with three sequential layers of a neural network that combine the outputs from all previously hidden layers. These dense layers outputs 104 feature vectors from Bert embeddings and 104 from syntactic and semantic were then concatenated together and passed through final dense layers.The output was then passed through a sigmoid activation function. The model was trained with pre-trained Colbert parameters using our feature modifications. Adam optimizer and Binary Cross entropy loss were used for training and the modified model was trained for 10 epochs with a batch size of 64. These final layers aim to determine sentence congruity and detect changes in the reader's viewpoint after reading the punchline.
   \newline

\section{Alternatives and Justification}

Our proposed method aims to assess whether a sentence is humorous or not by considering three types of information within it. While there exist traditional methods such as SVM \cite{inproceedings} and XGBoost,\cite{chen2016xgboost}and simpler sequence-to-sequence models like LSTM-based ones\cite{patel2021laugh}, the complexity of humor detection may challenge their ability to understand the full context. In contrast, more advanced models like BERT can handle this complexity better. There can also be graph-based approaches to find the relationship between sentences and words before training. 

BERT \cite{devlin2018bert} distinguishes different meanings for particular words based on the context by using contextualized embeddings. For example, even though the word “stick” could be both a noun as well as a verb, normal word embeddings assign the same vector to both meanings. BERT is trained in a self-supervised way by predicting missing words in sentences, and predicting if two randomly chosen sentences are subsequent or not.
Previously, there was an attempt to use BERT \cite{weller2019humor}\cite{devlin2018bert}for humor detection, and it worked well on the training data. However, it didn't consider syntactical information. The only result provided was its performance on a Spanish Tweet Dataset, where it performed significantly worse (about a 15\% drop).

\subsection{Challenges Encountered}
Training the COLBERT model was both interesting and challenging. The biggest hurdle was needing a lot of computer power, like high-performance GPUs. The training also took a long time because the model is complex. Making changes to COLBERT for specific tasks required tweaking its structure. These challenges taught us a lot about large language models and their model layer specifications.  It also showed that managing costs is crucial. Overall, the project highlighted how vital it is for models to be both powerful and flexible.

While creating features for ambiguity, we encountered several challenges. The ambiguity structure includes the sense combination score, farmost, and closest path similarity. During the process of determining path similarity using WordNet \cite{Fellbaum1998} , we grappled with utilizing the Synonym set for a given word. WordNet contains synsets for words in various forms like nouns, verbs, etc. Our objective was to exclusively identify noun meanings for a given noun, which posed a significant challenge. Obtaining information about path similarity turned out to be a time-consuming process. Programming the formulas mentioned in previous literature also presented a challenging task. Surprisingly, we anticipated that semantic features would yield higher accuracy compared to syntactic features, given our focus on understanding the meanings of words and sentences; however, the opposite turned out to be true.

{We believe that combining syntactical information with contextual and semantic information can lead to better performance on unseen data. Humor often involves a mix of syntactic cues, semantic meaning, and context, and considering all these aspects should improve the model's accuracy in detecting humor.}

\section{Evaluation}
To evaluate the effectiveness of our approach for humor detection we employed a range of evaluation metrics. Accuracy, F1 score, and ROC-AUC are fundamental metrics to gauge the model's overall performance, its balance between precision and recall, and its ability to distinguish between humorous and non-humorous instances. During training we employed different methodologies, for all the models not involving BERT in any way, the data was split into train and val, and the hyperparameters were set using the val data. For BERT the whole dataset was used to finetune.
To evaluate its performance on unseen data, we scraped data from a website \cite{bestlife2020oneliners} and also used a sample of jokes scraped from Reddit by a third-party \cite{orionw2020reddithumordetection}.

\subsection{Results}
We primarily assess our hand-crafted features based on accuracy and the F1 measure. The accuracy of the combined features, encompassing all three types, ranked the highest, followed by the syntactic features. These outcomes are illustrated in the table below. Evaluating BERT and Colbert models based on accuracy, we observed a consistent trend where the performance on unseen data was lower compared to the training data. These findings are also presented in the table below.

\begin{table}[t]
\centering
\begin{tabular}{|c|c|c|c|c|}
\hline
\textbf{Models} & \textbf{NRCLex} & \textbf{Syntactic} & \textbf{Semantic} & \textbf{Combined} \\
\hline
Decision Tree & 0.61 & 0.72 & 0.67 & \textbf{0.74}\\
\hline
Gradient Boost & 0.61 & 0.71 & 0.65 & \textbf{0.72}\\
\hline
\end{tabular}
\vspace{5pt}
\caption{Train data Results on hand-crafted features of 3 types and combined}
label{table:results}
\end{table}

\begin{table}[t]
\centering
\begin{tabular}{|c|l|l|l|l|}
\hline
\textbf{Model} & \multicolumn{2}{c|}{Without Features} & \multicolumn{2}{c|}{With Features} \\
\hline
Model & Accuracy & F1 & Accuracy & F1\\
\hline
BERT & 0.50 & 0.37 & 0.52 & 0.39\\
\hline
ColBERT & 0.50 & 0.34 & \textbf{0.62} & 0.60\\
\hline
\end{tabular}
\vspace{5pt}
\caption{Results on Test (Reddit) dataset with and without features.}
label{table:results2}
\end{table}

\begin{figure}[htbp]
\centering
    \includegraphics[width=0.45\textwidth]{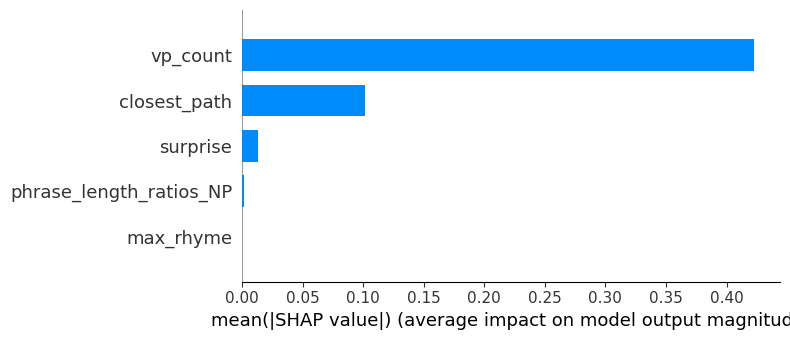}
    \caption{SHAP on all 33 features}
    \label{fig:shap1}
\end{figure}

\begin{figure}[ht]
    \centering
    \includegraphics[width=0.5\textwidth]{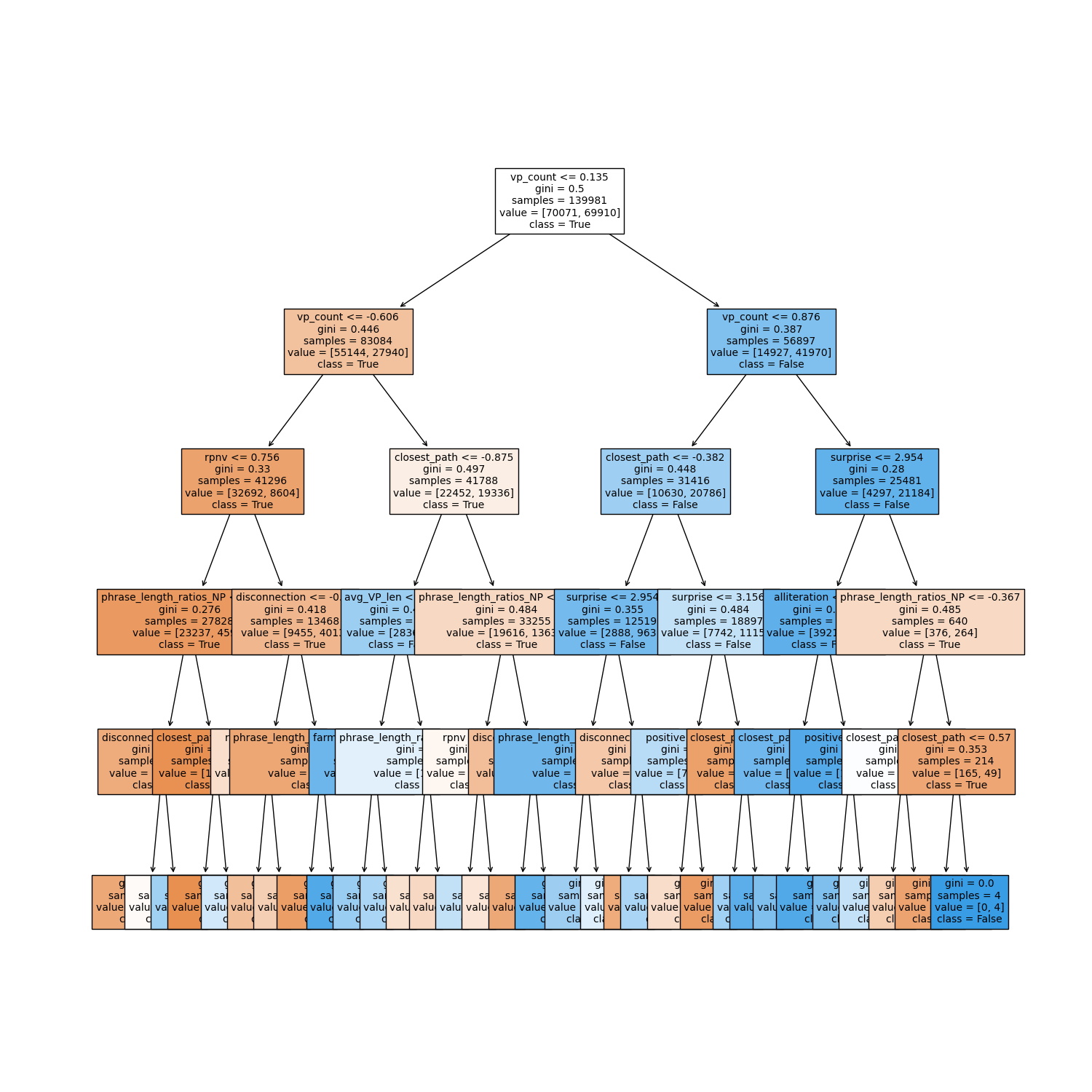}
    \caption{DT on all 33 features}
    \label{fig:dt}
\end{figure}
\section{Main Findings}
We've discovered intriguing nuances in how features representing emotions and semantics help the models in humor detection. Looking at the SHAP and Decision trees it was clear that it is easier for machines to detect humor based on words that exhibit emotions such as positive, anticipation, surprise, and trust. It is quite intriguing to see anticipation and surprise as they are two fundamental features that affect whether something is funny or not. 
Upon looking at the statistics of the syntactical elements, it seems VP\_count is one feature that helps the models to understand if a sentence would be funny or not, and even after combining all the 33 features and feeding it into the model, it seems that VP\_count is still what provides more understanding the model. From the features obtained from NRCLex, only the \textit{surprise} feature seems to be present in the top 5 features, which is surprising as we expected anticipation to be present as well. But what wasn't surprising was the domination of semantic features over syntactic, which is quite understandable as you cannot take something at its face value, just because a sentence contains syntax that might contain emotions projecting humor, the whole sentence need not be humorous, compared to a sentence having a funny meaning is highly probable to be funny.
But on standalone terms, embeddings that capture the contextual information, and are produced by models that are pre-trained on a huge corpus dominate every other feature. Does it mean this experimentation was a waste of time? No. Combining the above-mentioned features with the embeddings obtained from models such as BERT, resulted in higher quality which performed better on the downstream classification task, this can be seen even in the results.
Overall there is comparable significance of adding syntactical and semantic elements to the contextual meanings. 

\section{Limitations}

\subsection{Absence of Audio Data}
The task of humor classification, especially when dealing with textual data, is a complex and challenging task due to several factors and One of the main limitations in this project is not having audio along with text. A sentence can have multiple interpretations, some interpretations might be humorous while some might not.

For example, take the sentence: "Did you hear about the guy whose whole left side was cut off? He's all right now." The humor here relies on how you say it, and if it's not said right, it might not be funny and could even sound insensitive. This kind of ambiguity is tough for people to navigate, and it's even harder for a  model.

Audio can capture things like tone and pronunciation, making it easier for the model to understand and recognize humor. So, adding audio might just help our model do a better job at figuring out what's funny.

\subsection{Feature Selection of Hand Crafted Features}

Out of several features mentioned in \cite{liu2018exploiting}\cite{yang2015humor}we finalized to work with only 33 features, while more features don't mean better performance, we can't help but wonder if they might perform better than the ones selected.

When we performed feature selection on the 33 features, we used SHAP and Decision Tree Classifier, while both are frequently used practices, further experimentation can be done with methods such as Correlation Analysis, Recursive Feature elimination, Recursive Feature addition, etc. 

\subsection{Bias}

Our model is trained on just a single dataset, while it is a collection of one-liners, they all have a similar pattern. While it does perform equally well on unseen data, we believe it is beneficial to train on multiple datasets, and on different types of humor for better generalization.

\subsection{Problem Formulation}

Our primary goal is to predict humor based on a given sentence. However, it's important to note that humor can sometimes be offensive or insensitive, and our current model might not recognize this aspect. It is crucial for the model to promptly identify offensive jokes and flag them. This proactive approach is intended to prevent any negative repercussions or backlash that may arise from the use of inappropriate content.

\subsection{Explainabilty}

The hand-crafted features can be used to explain part of the model's high performance, but the inherent black-box nature of the BERT/ANNs is still present as the contextual features cannot be explained to stakeholders properly why the model is succeeding/failing.

\section{Prerequisites to Apply in Practice}

We have made the entire codebase open source and ready to use. The code is hosted and available to use on GitHub.\footnote {\url{https://github.com/psvkaushik/LOLgarithm}}.

The only requirement for it to be able to work in practice is that the jokes should be in a text-like format to feed into the model. This is because, all the features are extracted from the sentence, and there is no pre-processing required to be done on the user's side.

\section{Future Work and Improvements}
Despite incorporating a range of one-liners from our dataset, the model's performance fell short of our expectations for unseen data. Even after experimenting with fine-tuning the model, ensuring the robustness of new data remained a challenge. This issue could stem from potential overfitting, especially for the BERT and Colbert models, which might not adequately handle structures beyond their design. One potential solution is to amalgamate multiple datasets and reassess the model's performance. Developing nuanced datasets could be instrumental in addressing this challenge. Additionally, exploring larger models like GPT \cite{radford2018improving} could offer an alternative approach.

\section{Conclusion}
In this study, we tested how well a model performs when we combine hand-crafted features with contextual embeddings. By examining jokes, we created features based on sentence structure and meaning, discovering that these features can enhance the model's ability to recognize humor. Our analysis revealed that humorous texts often: 1) use straightforward words in complex sentence structures, 2) include vivid elements like adverbs, and rhymes 3) relate closely in meaning within sentences, and 4) combine different word senses. These findings highlight the specific style elements found in humor.



%



\end{document}